# HyperImageNet: A Large-Scale High-Spatial-Resolution Hyperspectral Imagery Classification Benchmark

Chuguang Zeng
*Wuhan University*
*430072 Wuhan, China*
zcg2025@whu.edu.cn

Jingtao Li
*Wuhan University*
*430072 Wuhan, China*
JingtaoLi@whu.edu.cn

Yinhe Liu
*Wuhan University*
*430072 Wuhan, China*
liuyinhe@whu.edu.cn

Yanfei Zhong
*Wuhan University*
*430072 Wuhan, China*
zhongyanfei@whu.edu.cn

***Abstract*—We present HyperImageNet, a large-scale benchmark for fine-grained hyperspectral land-cover understanding. The dataset contains 26,084 airborne hyperspectral image patches with 224 spectral bands and 138 fine-grained land-cover categories. Unlike existing datasets, HyperImageNet provides raw imagery, pixel-level semantic labels, and object-level instance masks, supporting both semantic and instance segmentation. Furthermore, we establish an open-environment benchmark with strict spatial separation to evaluate representative methods and the HyperFree foundation model. Experimental results demonstrate the effectiveness of HyperImageNet for fine-grained hyperspectral understanding and open-environment remote sensing research.**

## I. Introduction

Hyperspectral remote sensing imagery (HSI), characterized by continuous and fine-grained spectral features, possesses significant advantages in fields such as precise land-cover classification, environmental monitoring, and resource exploration. However, compared to the large-scale open-source datasets prevalent in the general computer vision domain (e.g., RGB images), high-quality hyperspectral datasets remain exceedingly scarce. This scarcity primarily stems from the high acquisition costs of HSI, constraints on spatial resolution, and the labor-intensive annotation process that heavily relies on expert knowledge. Consequently, this data shortage has become a critical realistic bottleneck hindering the evolution of hyperspectral deep learning models, particularly large-scale foundation models. To address this pragmatic dilemma, contemporary research primarily follows three avenues: (1) continuously conducting large-scale data collection and open-sourcing efforts, such as HySpecNet-11k[1]; (2) utilizing generative models to synthesize simulated data, such as HSIGene[2], though simulated data often fails to fully reconstruct the complex physical degradation processes inherent in ground spectral mechanisms, thereby limiting its generalizability; and (3) introducing pretrained foundation models with powerful high-dimensional feature extraction and generalization capabilities, such as SpectralEarth-FM[3], to more efficiently exploit the potential of existing hyperspectral imagery.

In the exploration of foundation model applications, the recently proposed HyperFree[4] framework tailored for HSI has demonstrated outstanding zero-shot segmentation performance. Nevertheless, while such models can generate high-quality geometric segmentation masks from unlabeled data, their outputs are typically semantic-agnostic. This decoupling between geometric boundaries and explicit semantics severely restricts the profound application of hyperspectral foundation models in specific scene-level tasks. To bridge this semantic perception gap, we construct a large-scale $H^2$(high spatial and high spectral resolution) benchmark dataset named HyperImageNet. The raw imagery of this dataset is acquired by the AVIRIS-Classic and AVIRIS-NG airborne sensors, encompassing 224 continuous spectral bands with a spatial resolution ranging from 0.5 to 5 meters. Regarding the construction of the labeling system, we transcend the limitations of a single data source by seamlessly fusing the annual Cropland Data Layer (CDL) product from the United States Department of Agriculture (USDA) with massive crowdsourced geospatial data from Overture Maps. Consequently, we successfully filter and compile 26,084 high-quality image patches covering 138 fine-grained land-cover subcategories. Its ultra-fine-grained characteristics provide a solid data foundation for migrating cutting-edge land-cover recognition tasks, such as open-vocabulary classification, into the hyperspectral domain.

Distinct from conventional hyperspectral datasets that merely provide a single label modality, HyperImageNet pioneers a trinity data structure: raw $H^2$ imagery (TIF), pixel-level semantic labels (TIF), and object-level instance masks (COCO JSON) generated via the HyperFree framework followed by meticulous manual refinement and post-processing. This design simultaneously provides pixel-level semantic information and object-level instance boundaries for the same image, breaking through the monolithic task dimension of prior datasets and enabling concurrent support for multiple complex tasks, including semantic and instance segmentation. Furthermore, considering that previous hyperspectral tasks were often confined to a single-image split for training and testing sets—which induces severe data leakage problems—this paper selects 31 representative land-cover categories alongside an open unknown class containing multiple land-cover types. By enforcing strict physical and spatial isolation between the training and testing sets, we establish a more challenging open-environment evaluation protocol and complete preliminary benchmark testing through fine-tuning the HyperFree architecture. HyperImageNet not only verifies that multi-tier fine-grained classification datasets can effectively mitigate the semantic vacuum of foundation models but also serves as a novel reference for future explorations into ultra-fine-grained hyperspectral land-cover recognition algorithms.

## II. Related Work

The advancement of the hyperspectral remote sensing domain is inextricably linked to the driving force of early semantic segmentation datasets (summarized in Table I) [5]–[13]. However, constrained by data acquisition platforms and annotation costs, existing public datasets often struggle to balance data scale, spatial-spectral resolution, and annotation granularity.

TABLE I. COMPARISON OF HYPERSPECTRAL SEMANTIC SEGMENTATION DATASETS

| *Dataset* | *Spatial Resolution (m)* | *Spectral Range (nm)* | *Bands* | *No. of Images* | *No. of Classes* |
|---|---|---|---|---|---|
| Washington DC[5] | \ | 400-2400 | 191 | 2 | 7 |
| Pavia University[6] | 1.3 | 430-860 | 103 | 2 | 9 |
| Pavia Center[6] | 1.3 | 430-860 | 102 | 4 | 9 |
| Urban | 2 | 380-2500 | 210 | 1 | 4 |
| KSC[7] | 18 | 400-2500 | 176 | 2 | 13 |
| Botswana | 30 | 400-2500 | 145 | 3 | 14 |
| Indian Pines[8] | 20 | 400-2500 | 220 | 1 | 16 |
| Salinas Valley[8] | 3.7 | 370-2480 | 204 | 1 | 16 |
| Houston 2013[9] | 2.5 | 380-1050 | 144 | 4 | 15 |
| Houston 2018[10] | 1 | 380-1050 | 48 | 10 | 20 |
| XiongAn[11] | 0.5 | 400-1000 | 250 | 21 | 19 |
| Chikusei[12] | 2.5 | 343-1018 | 128 | 25 | 19 |
| WHU-Hi-LK[13] | 0.463 | 400-1000 | 270 | 1 | 9 |
| WHU-Hi-HC[13] | 0.109 | 400-1000 | 274 | 1 | 16 |
| WHU-Hi-HH[13] | 0.043 | 400-1000 | 270 | 1 | 22 |
| WHU-OHS[14] | 10 | 400-1000 | 32 | 7795 | 16 |
| **Hyper ImageNet (Ours)** | **0.5-5** | **380-2510** | **224** | **26084** | **138** |

As illustrated in Table I, current datasets primarily face two major limitations. The first limitation lies in the constrained scale of image data and the inherent challenge of simultaneously achieving $H^2$ characteristics. The vast majority of classic datasets possess extremely small data volumes, making it difficult to support the cross-scene generalization evaluation of deep learning models. Conversely, datasets that achieve a substantial scale or possess high spatial resolution typically restrict their spectral coverage to the Visible and Near-Infrared (VNIR) range, lacking the continuous spectral features of the Short-Wave Infrared (SWIR) bands.

The second limitation involves the inadequate granularity of classification labels. The total number of categories in mainstream datasets generally does not exceed 20, primarily targeting macroscopic land-cover types. This coarse categorization fails to fully leverage the inherent advantages of hyperspectral imagery for fine-grained land-cover classification.

In contrast, the proposed HyperImageNet dataset not only boasts a massive volume of 26,084 images but also successfully realizes the $H^2$ properties, balancing a spatial resolution of 0.5 to 5 meters with full-spectrum continuous coverage from 380 to 2510 nm (across 224 bands), featuring 138 fine-grained subcategory labels. Furthermore, to bridge the semantic gap of foundation models in scene-level applications, this dataset introduces a tripartite data structure comprising raw $H^2$ imagery, pixel-level semantic labels, and object-level instance masks. This paired design transcends the single-task paradigm of previous benchmark datasets, establishing a comprehensive novel benchmark for the fine-grained semantic understanding of hyperspectral scenes and the rigorous evaluation of foundation models.

## III. THE HYPERIMAGENET DATASET

Compared to spaceborne hyperspectral sensors, which are susceptible to atmospheric interference and limited in spatial resolution, airborne hyperspectral imaging platforms can provide ultra-high spatial resolution (ranging from sub-meter to meter scales) and continuous spectra with high signal-to-noise ratios while conducting large-scale observations. Such high-quality $H^2$ physical properties are prerequisites for achieving ultra-fine-grained land-cover parsing in real-world scenarios, which constitutes the core motivation for selecting an airborne platform to construct our dataset. However, building large-scale labels matching this level of precision is a highly challenging engineering endeavor. Although introducing existing authoritative land-cover products (e.g., USDA CDL) and open-source crowdsourced geospatial data (e.g., Overture Maps) enables the semi-automatic annotation of basic polygonal and point features, the pixel-level accuracy of automated labels cannot be guaranteed due to spatial offsets among multi-source data and the ambiguity of fine-grained semantics. Therefore, to ensure the reliability of the benchmark data, extensive manual verification and meticulous refinement were continuously invested after fusing the multi-source labels. This section details the entire construction pipeline of HyperImageNet, encompassing the acquisition and preprocessing of airborne hyperspectral imagery, the annotation strategy for the 138 multi-source fine-grained labels, and the generation mechanism of object-level masks.

### A. Data Acquisition and Preprocessing

The raw hyperspectral imagery of HyperImageNet is sourced from the Airborne Visible/Infrared Imaging Spectrometer (AVIRIS) platforms. To balance the observation time span and the overall scale of the data, this dataset comprises two image subsets: HyperSeg-A and HyperSeg-B.

Specifically, HyperSeg-A aggregates 22,699 image patches acquired by the AVIRIS-Classic sensor between 2009 and 2022, natively providing 224 continuous spectral bands across the 380-2510 nm range. HyperSeg-B contains 3,385 image patches collected by the next-generation AVIRIS-NG sensor in 2023, featuring up to 425 native spectral bands. Considering the strict requirements of deep learning models regarding input feature dimensions, we implemented a cross-sensor spectral alignment strategy during the preprocessing stage to guarantee consistency in the spectral feature space across the entire dataset. Based on the principle of central wavelength matching, we extracted 224 bands from the 425 bands of AVIRIS-NG that highly correspond to those of AVIRIS-Classic. Through unified preprocessing, these 26,084 images spanning different years and sensors were successfully aligned in their spectral bands, collaboratively building an $H^2$ base imagery library with both a large time span and high spatial resolutions of 0.5 to 5 meters.

### B. *Fine-Grained Labeling Strategy*

The core challenge in constructing the 138 fine-grained subcategory labels lies in the fact that a single data source can hardly maintain extremely high semantic granularity simultaneously across vast natural landscapes and complex urban scenes. Consequently, we adopted a differentiated multi-source label fusion strategy based on the image acquisition years: for the HyperSeg-A subset, which lacks the support of early high-precision crowdsourced data, the annual USDA CDL raster product serves as the label foundation; for the HyperSeg-B subset, Overture Maps full-pixel sampling point data is additionally introduced to establish a semantically complementary annotation system.

Regarding multi-source data fusion and category establishment, the CDL and Overture Maps form a complementary advantage based on the physical features of land covers. The CDL data exhibits extreme professionalism in classifying natural and agricultural vegetation such as crop subcategories. This study fully retains these fine-grained classifications, which possess significant potential for hyperspectral discrimination. However, the label differentiation of urban scenes in CDL, based on Development Intensity, often lacks significant variations in hyperspectral features. To address this limitation, we initially merged these categories weakly correlated with hyperspectral tasks, and then introduced Overture Maps sampling point data to semantically refine the reconstructed category labels. Abundant in crowdsourced geographic information, Overture Maps can provide higher-quality micro-category annotations in urban scenes (e.g., specific roof materials and subdivided road networks), effectively filling the gap of CDL concerning urban hyperspectral properties. During the system construction, we strictly implemented the core principle of "fine-granularity first." To ensure the evaluation potential of hyperspectral data in cutting-edge tasks like long-tail distribution and rare land-cover recognition, we insisted on retaining categories with extremely low percentages of valid pixels as independent classes.

Regarding the spatial alignment mechanism of multi-source labels, benefiting from the orthorectified raw imagery and data information tables provided by the AVIRIS platform, we first performed matching via a latitude-longitude grid and subsequently mapped the JSON sampling points of Overture Maps to the corresponding raster pixels, thereby establishing an initial fine-grained label library covering the entire scene. Ultimately, following meticulous manual review of the single-class labels for each image, 138 fine-grained subcategory labels highly consistent with hyperspectral perception mechanisms were established.

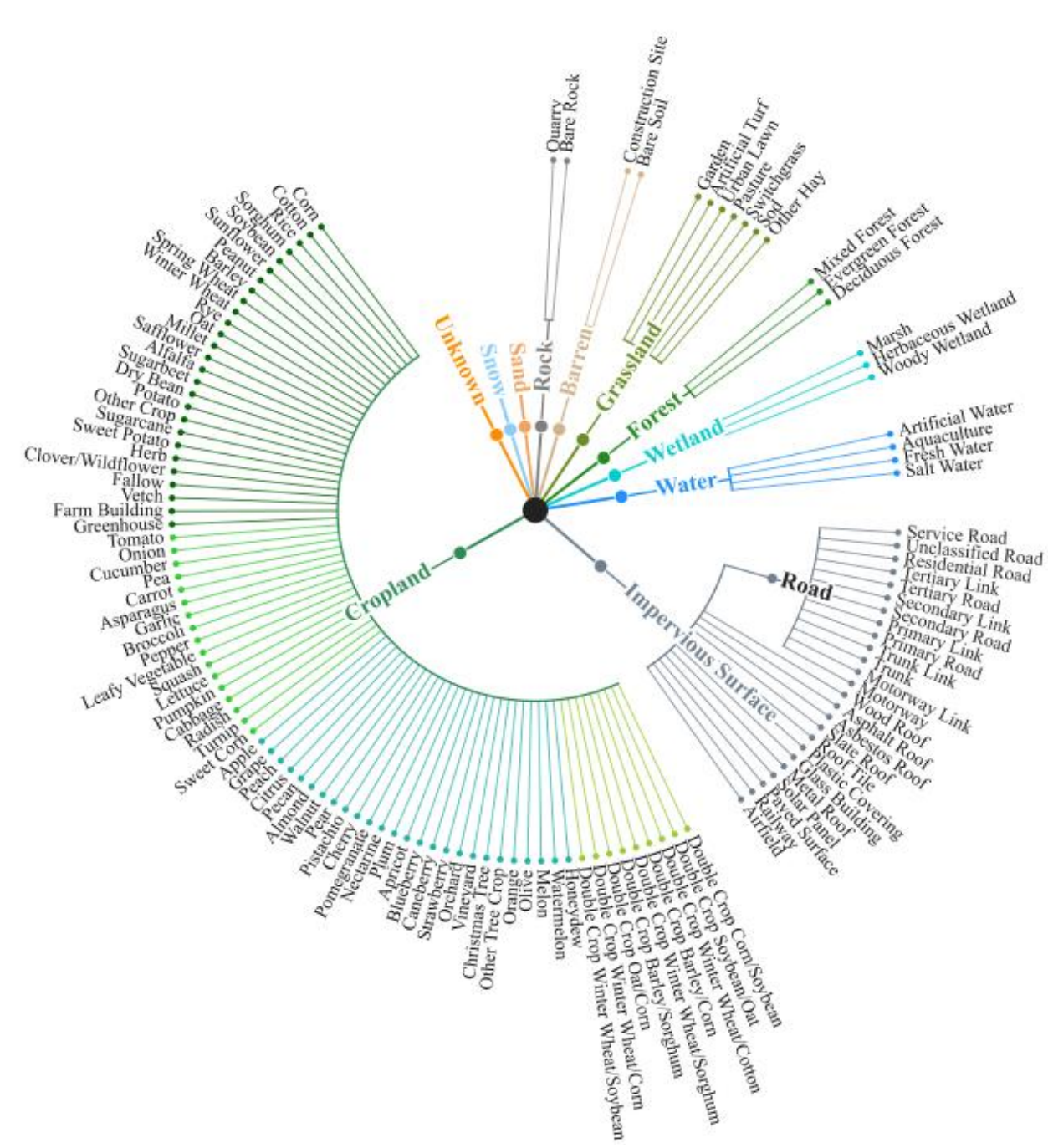

Fig. 1. The hierarchical classification system of HyperImagenet

As shown in Fig. 1, to systematically manage these 138 ultra-fine-grained categories, HyperImageNet establishes a rigorous hierarchical land-cover taxonomy. At the macro level, the dataset comprises 11 major categories: cropland, impervious surfaces, forests, shrublands, grasslands, wetlands, water bodies, barren land, and snow. At the subcategory level, it demonstrates extreme potential for granular recognition, encompassing not only over 80 specific crop species (e.g., cucumbers, tomatoes, strawberries) and tree components of different phenological traits (e.g., evergreen forests, mixed forests), but also precise architectural materials (e.g., solar panels, metal roofs), artificial facilities (e.g., artificial turf), and distinctions in water properties (e.g., freshwater and saltwater). This taxonomic architecture not only preserves the semantic coherence of macro-ecology but also effectively leverages the natural spectral advantages of hyperspectral data in fine-grained land-cover classification.

### C. *Object-Level Mask Generation*

To construct the aforementioned tripartite data structure of "raw imagery - category labels - segmentation masks" and subsequently bridge the semantic gap of foundation models in scene applications, this paper employed the HyperFree framework to generate object-level segmentation masks aligned with the pixel-level labels. Given that HyperFree possesses a promptable segmentation architecture similar to the Segment Anything Model (SAM)[15], we systematically fine-tuned its core hyperparameters tailored to the complex spatial-spectral characteristics of AVIRIS imagery. Specifically, through iterative scene-adaptive testing, we primarily optimized the model's Prediction IoU Threshold, Stability Score Threshold, and the feature input range for specific bands, thereby obtaining the preliminary segmentation masks.

However, to capture fine-grained land covers as comprehensively as possible, the relatively low prediction thresholds set in the initial processing inevitably induced a massive adjacent adhesion phenomenon among small-area masks. To tackle this issue, we introduced a targeted morphological post-processing mechanism. First, a morphological opening operation was applied to effectively detach the adhered mask boundaries; subsequently, precise filtering and cleaning were conducted in conjunction with connected component area thresholds tailored to the characteristics of fragmented isolated noise pixels. Refined through this pipeline, the final generated COCO-format masks effectively eliminated noise while preserving high-precision geometric boundary demarcations. An illustrative visualization of the tripartite data is presented in Fig. 2 below.

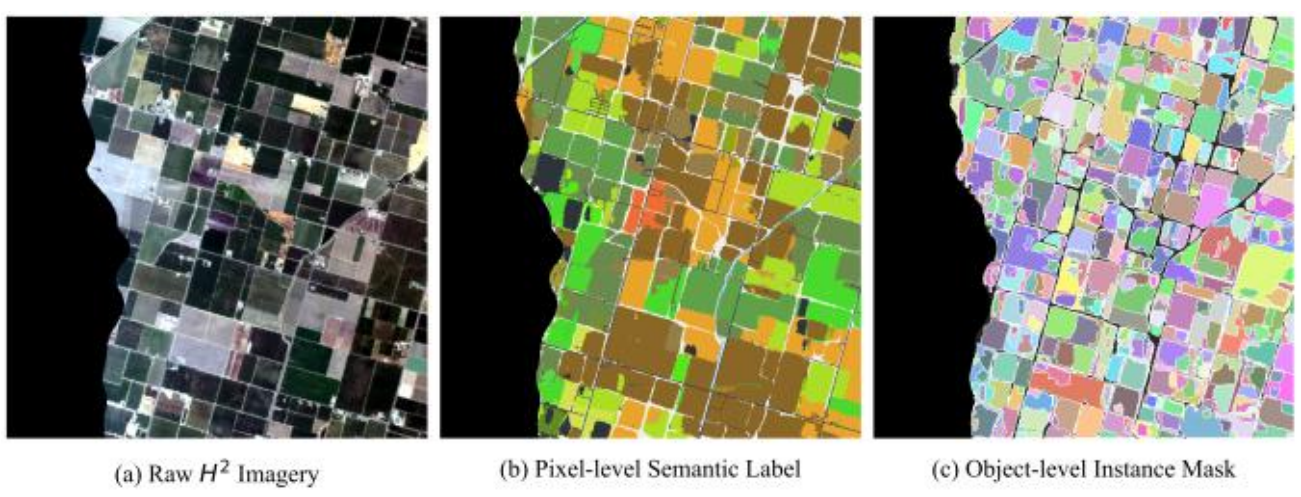

Fig. 2. Raw imagery - category labels - segmentation masks example

## IV. BENCHMARK EVALUATION

To verify the application potential of HyperImageNet in real-world complex scenarios and systematically evaluate the performance boundaries of existing hyperspectral segmentation algorithms, we extracted a high-quality subset from the constructed dataset to design the following benchmark. Distinct from the common single-image splitting strategy in traditional hyperspectral classification tasks, this benchmark strictly adheres to the principle of spatial physical isolation to thoroughly eliminate the Data Leakage problem caused by spatial contexts during model training.

### A. *Experimental Setup and Dataset Partitioning*

From the vast image library of HyperImageNet, we meticulously selected 562 high-quality $H^2$ (high spatial and high spectral resolution) image patches characterized by highly complex scenes and multi-tier landform features as the benchmark subset. This subset retains the original 224 spectral bands and spatial resolutions ranging from 0.5 to 5 meters.

TABLE II. STATISTICAL INFORMATION OF THE $H^2$ FINE-GRAINED CLASSIFICATION DATASET

| ID | Class | No. of Train Images | No. of Train Pixels | No. of Test Images | No. of Test Pixels |
|---|---|---|---|---|---|
| 0 | Unlabeled | / | / | / | / |
| 1 | Water | 28 | 220,328 | 12 | 82,254 |
| 2 | Impervious Surface | 28 | 301,412 | 12 | 143,123 |
| 3 | Barren | 7 | 720,917 | 3 | 41,151 |
| 4 | Snow | 8 | 247,846 | 3 | 101,209 |
| 5 | Woody Wetland | 49 | 2,620,751 | 21 | 638,403 |
| 6 | Herbaceous Wetland | 41 | 2,555,855 | 17 | 1,075,456 |
| 7 | Grassland | 46 | 2,563,866 | 20 | 1,063,724 |
| 8 | Other Hay | 19 | 300,771 | 7 | 114,024 |
| 9 | Deciduous Forest | 52 | 3,239,035 | 22 | 1,159,919 |
| 10 | Evergreen Forest | 49 | 3,026,683 | 21 | 1,112,395 |
| 11 | Mixed Forest | 40 | 2,016,434 | 17 | 726,688 |
| 12 | Shrubland | 24 | 2,043,253 | 10 | 846,451 |
| 13 | Corn | 12 | 733,831 | 5 | 371,827 |
| 14 | Cotton | 11 | 283,399 | 4 | 41,455 |
| 15 | Soybean | 9 | 349,742 | 4 | 141,073 |
| 16 | Sunflower | 7 | 115,121 | 3 | 63,639 |
| 17 | Barley | 7 | 87,516 | 3 | 69,061 |
| 18 | Winter Wheat | 9 | 455,581 | 4 | 265,808 |
| 19 | Alfalfa | 21 | 978,140 | 9 | 319,994 |
| 20 | Dry Bean | 20 | 133,581 | 9 | 90,960 |
| 21 | Tomato | 9 | 629,884 | 3 | 178,979 |
| 22 | Fallow | 43 | 2,030,482 | 16 | 743,842 |
| 23 | Strawberry | 8 | 63,243 | 3 | 35,567 |
| 24 | Peach | 8 | 1,042,367 | 3 | 283,520 |
| 25 | Grape | 27 | 1,514,226 | 14 | 666,061 |
| 26 | Citrus | 15 | 125,960 | 6 | 46,526 |
| 27 | Orange | 46 | 3,110,256 | 22 | 1,380,311 |
| 28 | Olive | 8 | 247,507 | 3 | 84,796 |
| 29 | Almond | 39 | 2,370,228 | 16 | 969,074 |
| 30 | Walnut | 10 | 502,412 | 3 | 105,516 |
| 31 | Pistachio | 11 | 711,910 | 5 | 367,959 |
| 32 | Unknown | / | / | 84 | 623,449 |

As shown in Table II, regarding the category system and data partitioning, this benchmark constructs an open-environment evaluation framework comprising 31 known classes + 1 unknown class. The 31 known classes cover water bodies, impervious surfaces, barren land, snow, woody wetlands, herbaceous wetlands, grasslands, fallow land, and over 20 fine-grained agricultural and forestry crops (such as typical categories like corn, evergreen forests, grapes, and pistachios). The 32nd class serves as the "Unknown" category in the open environment. To ensure the absolute objectivity of the open-set evaluation, this category remains completely invisible to the models during the training phase, implicitly encompassing 21 specific unseen subcategories (e.g., specific types of rice, etc.).

To guarantee the fairness and generalization reliability of the evaluation results, the 562 images are strictly divided into a training set (383 images) and a testing set (179 images) at a ratio of approximately 7:3. The training and testing sets are completely independent in geographic space, demanding that the models possess genuine generalization capabilities under the interference of unknown spatial regions and unseen land covers.

### B. *Closed-Set Binary Segmentation Benchmark*

To comprehensively evaluate the feature extraction limits of mainstream deep learning models on singular hyperspectral land covers, and to verify the dataset's efficacy in bridging the semantic gap of foundation models, we first constructed an independent One-vs-Rest Binary Segmentation benchmark targeting the 31 known classes. In this track, we selected 6 representative baseline methods for comparison, including classic Convolutional Neural Networks (CNN[16], UNet[17]), a hybrid model combining spatial-spectral features (HybridSN[18]), a Transformer-based architecture (TransUNet[19]), a deep network explicitly designed for full-image hyperspectral semantic segmentation (FreeNet[20]), and a hyperspectral foundation model (our fine-tuned HyperFree). The evaluation metrics employed are the F1-score for each category and the mean F1-score (mF1).

TABLE III. F1-SCORE COMPARISONS OF THE 6 METHODS ACROSS THE 31 FINE-GRAINED CATEGORIES

| *ID* | *CNN* | *UNet* | *Free Net* | *Hybrid SN* | *Trans UNet* | *Hyper Free (Ours)* |
|---|---|---|---|---|---|---|
| 1 | 0.8513 | 0.8559 | 0.8596 | 0.8541 | **0.8488** | **0.9054** |
| 2 | 0.3468 | 0.4058 | 0.3698 | 0.4169 | **0.3005** | **0.4293** |
| 3 | **0.2529** | **0.6523** | 0.6339 | 0.2851 | 0.5165 | 0.3375 |
| 4 | **0.4450** | 0.6406 | 0. 5564 | 0.5108 | 0.6426 | **0.6525** |
| 5 | **0.5354** | 0.5655 | 0.5665 | 0.5667 | 0.5569 | **0.5776** |
| 6 | **0.4360** | 0. 4373 | 0.4635 | 0.4388 | 0.4787 | **0.4898** |
| 7 | **0.0000** | 0.5820 | 0.5826 | 0.5820 | **0.6284** | 0.5852 |
| 8 | **0.2523** | 0.4330 | 0.3328 | 0.3991 | 0.4251 | **0.4655** |
| 9 | **0.5078** | 0.5932 | 0.5776 | 0.5355 | 0.5503 | **0.5957** |
| 10 | **0.5633** | 0.5945 | 0.5899 | 0.5871 | 0.5929 | **0.5975** |
| 11 | **0.4041** | 0.4111 | 0.4188 | 0.4170 | 0.4176 | **0.4201** |
| 12 | **0.6647** | 0.6786 | 0.6746 | 0.6763 | 0.6782 | **0.6917** |
| 13 | **0.5230** | 0.5274 | **0.6351** | 0.5372 | 0.5329 | 0.5802 |
| 14 | 0.2935 | 0.3362 | **0.6372** | **0.2374** | 0.3846 | 0.4221 |
| 15 | 0.4318 | **0.4312** | 0.5004 | 0.5041 | **0.5820** | 0.5222 |
| 16 | **0.5275** | 0.6500 | 0.6026 | 0.6478 | 0.6413 | **0.6916** |
| 17 | 0.4293 | 0.5146 | **0.5539** | **0.4072** | 0.4959 | 0.5264 |
| 18 | **0.3957** | 0.4465 | **0.4907** | 0.4153 | 0.4236 | 0.4469 |
| 19 | 0.4099 | 0.4220 | **0.5055** | **0.3990** | 0.4816 | 0.4677 |
| 20 | **0.0000** | 0.2354 | 0.1740 | 0.2403 | 0.2313 | **0.2419** |
| 21 | **0.6602** | 0.7401 | 0.7401 | 0.6923 | 0.7348 | **0.7423** |
| 22 | 0.4536 | 0.4758 | **0.5769** | 0.4604 | **0.4501** | 0.4982 |
| 23 | **0.0450** | 0.2621 | 0.1729 | 0.1908 | 0.2538 | **0.2884** |
| 24 | **0.5466** | 0.5655 | **0.7055** | 0.5502 | 0.5486 | 0.5584 |
| 25 | **0.3251** | 0.3299 | **0.5590** | 0.4269 | 0.3620 | 0. 4295 |
| 26 | 0.0005 | **0.0000** | 0.1013 | 0.0031 | **0.0000** | 0.2853 |
| 27 | **0.5755** | 0.5954 | **0.6308** | 0.5853 | 0.5945 | 0.6069 |
| 28 | **0.4288** | 0.4753 | 0.4995 | 0.4388 | 0.5262 | **0.5632** |
| 29 | 0.5677 | 0.5825 | **0.4708** | 0.5360 | 0.5665 | **0.5876** |
| 30 | **0.4515** | 0.5123 | 0.4653 | 0.4829 | 0.5065 | **0.5137** |
| 31 | **0.6635** | 0.7091 | 0.6678 | 0.7171 | 0.6806 | **0.7555** |

Table III details the F1-score comparisons of the 6 methods across the 31 fine-grained categories. To intuitively display the performance hierarchy of each model, the table applies a red background to the highest accuracy in each column (i.e., each category), a yellow background for the second-highest, and a blue background for the lowest.

From the quantitative results, it is clearly observable that the HyperFree model demonstrates an overwhelming advantage, achieving optimal (red) or suboptimal (yellow) segmentation accuracies in the vast majority of categories (e.g., water bodies, snow, herbaceous wetlands, soybeans) with no bottleneck categories performing at the bottom (blue). Particularly in the classification of fine-grained crops (e.g., winter wheat, barley) and forests of different phenological traits (e.g., evergreen forests, mixed forests) where spectral features are highly prone to confusion, benefiting from the powerful high-dimensional spectral feature representation capability of the foundation model, HyperFree significantly outperforms traditional closed-set architectures such as UNet, HybridSN, and FreeNet.

### C. *Semantic Segmentation Benchmark in Open Environments*

Real-world remote sensing application environments are frequently open and fraught with unknowns. However, traditional closed-set segmentation networks (such as the previously tested CNN, UNet, and TransUNet) solidify the output dimensions of their classification heads by design, naturally lacking the capability to handle the "Unknown Class." When confronted with the 32nd open land cover, they inevitably and forcibly misclassify it as one of the 31 known classes.

Benefiting from the Prototype Learning mechanism—which mimics visual open-vocabulary tasks introduced during the fine-tuning phase—the fine-tuned HyperFree framework emerges as the sole foundation model in this test capable of Open-Set Recognition in open environments. It must be particularly emphasized that, to objectively evaluate the true performance of the models during full-image inference, this track switches the task paradigm to a 32-class joint classification encompassing the unknown class. Under this paradigm, severe feature boundary squeezing exists among the categories. The model must precisely delineate the mutually exclusive boundaries of unknown features without sacrificing the recall rate of known classes. Its optimization space is fundamentally distinct from closed-set binary segmentation; thus, the absolute numerical values of the two are not directly comparable.

TABLE IV. PERFORMANCE COMPARISON (MF1) ACROSS CLOSED-SET AND OPEN-SET

| ***Methods*** | | mF1 |
|---|---|---|
| **Closed-Set (1-31)** | ***CNN*** | 0.4190 |
| | ***UNet*** | 0.4911 |
| | ***FreeNet*** | 0.4964 |
| | ***HybridSN*** | 0.4755 |
| | ***TransUNet*** | 0.5043 |
| | ***HyperFree(Ours)*** | **0.5176** |
| **Open Set (1-32)** | ***HyperFree(Ours)*** | 0. 3594 |

In this highly challenging open-environment evaluation, the HyperFree model, fine-tuned with only limited supervision signals, achieved a Macro F1-score of 35.94% on the 1-32 class joint classification task (detailed category metrics are presented in Table IV). This authentic baseline result—which harbors immense room for improvement—not only proves that HyperImageNet can substantially bridge the semantic gap of hyperspectral foundation models, but also profoundly reveals the immense challenge and future research potential inherent in the task of hyperspectral open-environment segmentation.

## V. CONCLUSION

To address the semantic perception gap of hyperspectral foundation models in complex scene applications and the lack of fine-grained benchmarks, this paper constructs HyperImageNet, a large-scale $H^2$ (high spatial and high spectral resolution) remote sensing dataset. This dataset not only encompasses 26,084 image patches but also makes a breakthrough by establishing a fine-grained label taxonomy covering 138 subcategories. By providing a tripartite data structure of "raw imagery - semantic labels - object masks," HyperImageNet effectively supports the integration of pixel-level semantics and object-level geometric topologies. Furthermore, open-environment benchmark evaluations based on strict spatial physical isolation demonstrate that current mainstream hyperspectral feature representation paradigms face genuine generalization challenges under ultra-fine-grained classification and the interference of unknown categories. The open-sourcing of this dataset aims to provide a solid data infrastructure for long-tail distribution learning, open-environment perception, and the evaluation of multi-task foundation models in the hyperspectral domain, thereby propelling the advancement of intelligent land-cover parsing technologies.